\documentclass[11pt]{article}
\usepackage[]{acl}
\usepackage{times}
\usepackage{latexsym}
\usepackage{amsmath}
\usepackage{amssymb}
\usepackage{bm}
\usepackage{float}   
\usepackage{xcolor}
\usepackage{graphicx}
\usepackage{booktabs}
\usepackage{algorithm}
\usepackage{algpseudocode}
\usepackage{multirow}
\definecolor{closer}{RGB}{15,110,86}
\definecolor{human}{RGB}{90,90,95}
\usepackage{etoc}
\usepackage[table]{xcolor}
\usepackage{booktabs}   
\definecolor{gold}{HTML}{C9A24B}
\definecolor{aqblue}{RGB}{24, 95, 165}
\definecolor{eqgreen}{RGB}{15, 110, 86}
\definecolor{swamber}{RGB}{186, 117, 23}
\definecolor{commentgray}{RGB}{120, 120, 120}
\algnewcommand{\SectionComment}[1]{\Statex \textcolor{commentgray}{\textit{$\triangleright$ #1}}}

\usepackage{graphicx} 
\title{MINT: Min-Selection Preference Distillation for Balanced Multi-Objective Alignment}
\author{
  Tony Tu\textsuperscript{1,2} \and Sayan Chakraborty\textsuperscript{2} \and Ruomeng Xu\textsuperscript{2} \and Tony Qin\textsuperscript{2} \and Austin Tian\textsuperscript{2} \\
  \textsuperscript{1}Georgia Institute of Technology \quad \textsuperscript{2}Zillow Group \\
  \texttt{ttu32@gatech.edu} \quad \texttt{Sayan341@yahoo.in} \\
  \texttt{\{ruomengx, tonyqi, austinti\}@zillowgroup.com}
}
\date{July 2026}
\begin{document}
\maketitle
\begin{abstract}
Aligning a language agent to several objectives at once is a persistent
failure mode of preference-based training: when objectives are combined
additively, optimization collapses onto whichever is cheapest to improve
and sacrifices the rest, so a support agent learns to sound warm while
giving no real help. The root issue is that an additive reward has no
notion of \emph{balance}. We introduce \textsc{Mint}
(\textbf{MIN}-selection preference dis\textbf{T}illation), a one-line
change to preference distillation: rather than ranking sampled candidates
by a weighted sum of rewards, we rank them by their \emph{weakest}
objective, distilling the best-balanced candidate over the most lopsided
one with an unchanged DPO objective. This is the $p\!\to\!-\infty$ limit
of a generalized-mean family spanning additive to worst-case selection.
Across cooperative emotional support and adversarial negotiation,
min-selection lifts both objectives while sharply cutting their
imbalance; on emotional support it raises the weaker axis from $0.37$ to
$0.64$ ($p<10^{-40}$), surpassing human experts and persisting across
full multi-turn rollouts. A turn-by-turn analysis yields our central
finding: min-selection corrects imbalance in proportion to how imbalanced
the reference policy is, and its benefit endures over an interaction
precisely as long as that imbalance does.
\end{abstract}
\section{Introduction}
\label{sec:intro}
Aligning language models to human preferences with reinforcement learning
from human feedback \citep{christiano2017deep, ouyang2022training} has
become the dominant recipe for turning capable base models into useful
assistants, and Direct Preference Optimization
\citep{rafailov2023direct} has made this alignment dramatically simpler by
collapsing reward modeling and policy optimization into a single
supervised objective on preference pairs. Yet real deployments rarely
optimize a single notion of ``good.'' A dialogue agent must be helpful
\emph{and} harmless, informative \emph{and} concise, effective \emph{and}
kind, and these objectives routinely conflict
\citep{bai2022training, rame2023rewarded, zhou2024beyond}. The standard
response is to \emph{scalarize}: combine the objectives into a weighted
sum and optimize the aggregate, whether by training against a linear
combination of reward models, folding weights into the DPO loss
\citep{zhou2024beyond}, or interpolating per-objective policies along the
Pareto front \citep{rame2023rewarded}. Additive scalarization, however,
carries a structural flaw. Because a large gain on one objective can
offset a large loss on another, the aggregate reward cannot distinguish a
response that is excellent on every axis from one that is dominant on the
cheapest axis and empty on the rest. Optimization therefore tends to
collapse onto whichever objective is easiest to improve, an ``alignment
tax'' in which balance is silently traded away
\citep{ouyang2022training, rame2023rewarded}, and the resulting agent
learns to sound warm while giving no real help, or to win the deal while
alienating the counterpart.

We argue that this failure is not inherent to preference distillation but
to the \emph{additive} aggregation it conventionally uses, and that a
different aggregation removes it. We introduce \textsc{Mint}
(min-selection preference distillation), which retains the now-standard
best-of-$K$ distillation pipeline, sample $K$ candidates, score each with
a frozen judge, and distill the preference back into the policy via DPO
\citep{dong2023raft, gulcehre2023reinforced, rafailov2023direct}, but
changes the one step that determines what the policy actually learns from:
the selection of the chosen and rejected responses. Instead of ranking
candidates by a scalar or a weighted sum, \textsc{Mint} ranks them by
their \emph{minimum} objective, a Chebyshev worst-case criterion that is
the $p\!\to\!-\infty$ limit of the generalized power mean. The chosen
response is the candidate whose weakest axis is strongest, and the
rejected response the most lopsided, so the distilled policy is pulled
toward outputs that are good on \emph{every} objective rather than
outstanding on one. The change is deliberately minimal: it touches neither
the DPO objective, the reward model, nor the sampling procedure, and so
composes with the broad family of preference-optimization variants
\citep{azar2024general, ethayarajh2023kto, hong2024orpo, zhao2023slic}.
Its contribution is empirical rather than algorithmic, showing that the
selection geometry alone suffices to induce multi-objective balance.

We evaluate \textsc{Mint} on two deliberately opposite dialogue domains
with distinct policy models, cooperative emotional support (ESConv) and
adversarial price negotiation (CraigslistBargain), each instrumented with
two orthogonal objectives scored by a frozen large-language-model judge:
an action quotient (AQ) for task progress and an emotional quotient (EQ)
for relational quality. Min-selection improves both objectives while
substantially reducing their imbalance, and where the supervised
reference is badly imbalanced it raises the weaker-axis reward from
$0.37$ to $0.64$, yielding responses more balanced than those of human
experts. Beyond the headline gain, a turn-resolved analysis of interactive
rollouts surfaces our central and, we argue, generalizable observation:
min-selection corrects objective imbalance in proportion to how imbalanced
the reference policy is, and its advantage compounds over a multi-turn
interaction exactly to the extent that the reference remains imbalanced
along the trajectory. This reframes min-selection not as a universal boost
but as a targeted, well-characterized remedy for the multi-objective
collapse that additive preferences invite. Our contributions are: (i) a
generalized-mean view of preference-pair selection that exposes a
continuous family of rules from additive to worst-case; (ii) \textsc{Mint},
its min-selection limit, as a drop-in modification to best-of-$K$
preference distillation; and (iii) an empirical characterization, across
two structurally distinct domains and both static and interactive
evaluation, of when and for how long balanced selection helps.
\section{Related Work}
\label{sec:related}

\paragraph{Preference optimization.}
Reinforcement learning from human feedback \citep{christiano2017deep,
ouyang2022training} aligns language models to human preferences by
training a reward model and optimizing the policy against it with PPO
\citep{schulman2017proximal}. Direct Preference Optimization
\citep{rafailov2023direct} reformulates this pipeline as a single
supervised objective on preference pairs, showing that the language
model is implicitly its own reward model and thereby avoiding an
explicit reward-modeling and RL stage. Its simplicity has produced a
large family of variants that modify the loss or the reference term,
including IPO \citep{azar2024general}, KTO \citep{ethayarajh2023kto},
ORPO \citep{hong2024orpo}, and SLiC-HF \citep{zhao2023slic}. Our method
is deliberately agnostic to these choices. \textsc{Mint} changes only
\emph{which} candidates become the chosen and rejected responses, and
uses the standard DPO loss unchanged, so it remains compatible with any
of these variants.

\paragraph{Multi-objective alignment.}
Real deployments require balancing several, often conflicting,
objectives, such as helpfulness versus harmlessness or informativeness
versus conciseness, for which no single reward suffices
\citep{rame2023rewarded, zhou2024beyond}. Most approaches scalarize the
objectives with a \emph{weighted sum}: multi-objective RLHF trains a
policy against a linear combination of reward models, MODPO
\citep{zhou2024beyond} folds objective weights into the DPO loss, and
Rewarded Soups \citep{rame2023rewarded} trains one policy per objective
and interpolates their weights to trace the Pareto front.
Rewards-in-Context \citep{yang2024rewards} instead conditions a single
model on target reward values supplied in the prompt. A recurring
difficulty, noted across this line of work, is that additive
scalarization permits an ``alignment tax'' in which one objective is
improved at the expense of another. \textsc{Mint} departs from the
additive convention. Rather than weighting objectives, it selects
preference pairs by the \emph{minimum} objective (a Chebyshev
criterion), which we show empirically drives the policy toward balanced
improvement rather than trading one axis for the other, and requires no
per-objective weights or auxiliary models.

\paragraph{Rejection sampling and best-of-$K$ distillation.}
A complementary line of work improves a policy by sampling multiple
candidates, scoring them, and fine-tuning on the best ones. RAFT
\citep{dong2023raft} and ReST \citep{gulcehre2023reinforced} perform
reward-ranked (rejection-sampling) fine-tuning, distilling a
best-of-$K$ selection back into the policy, and statistical rejection
sampling \citep{liu2024statistical} connects this selection to
preference optimization. Best-of-$N$ sampling itself is a strong
inference-time baseline \citep{amini2025variational}, and BOND
\citep{sessa2024bond} distills its behavior into a single model to avoid
the inference-time cost. \textsc{Mint} is an instance of this
best-of-$K$ preference-distillation paradigm. It generates $K$
candidates, scores them with a frozen judge, and distills the result via
DPO, but whereas prior work ranks candidates by a \emph{single} scalar
reward, \textsc{Mint} ranks them by the minimum over \emph{multiple}
objectives, making the selection geometry itself the mechanism for
multi-objective balance. Our contribution is thus empirical rather than
algorithmic: the same distillation machinery, with a different selection
rule, yields balanced multi-objective policies.
\section{Method}
\label{sec:method}

\begin{figure*}[t]
  \centering
  \includegraphics[width=\textwidth]{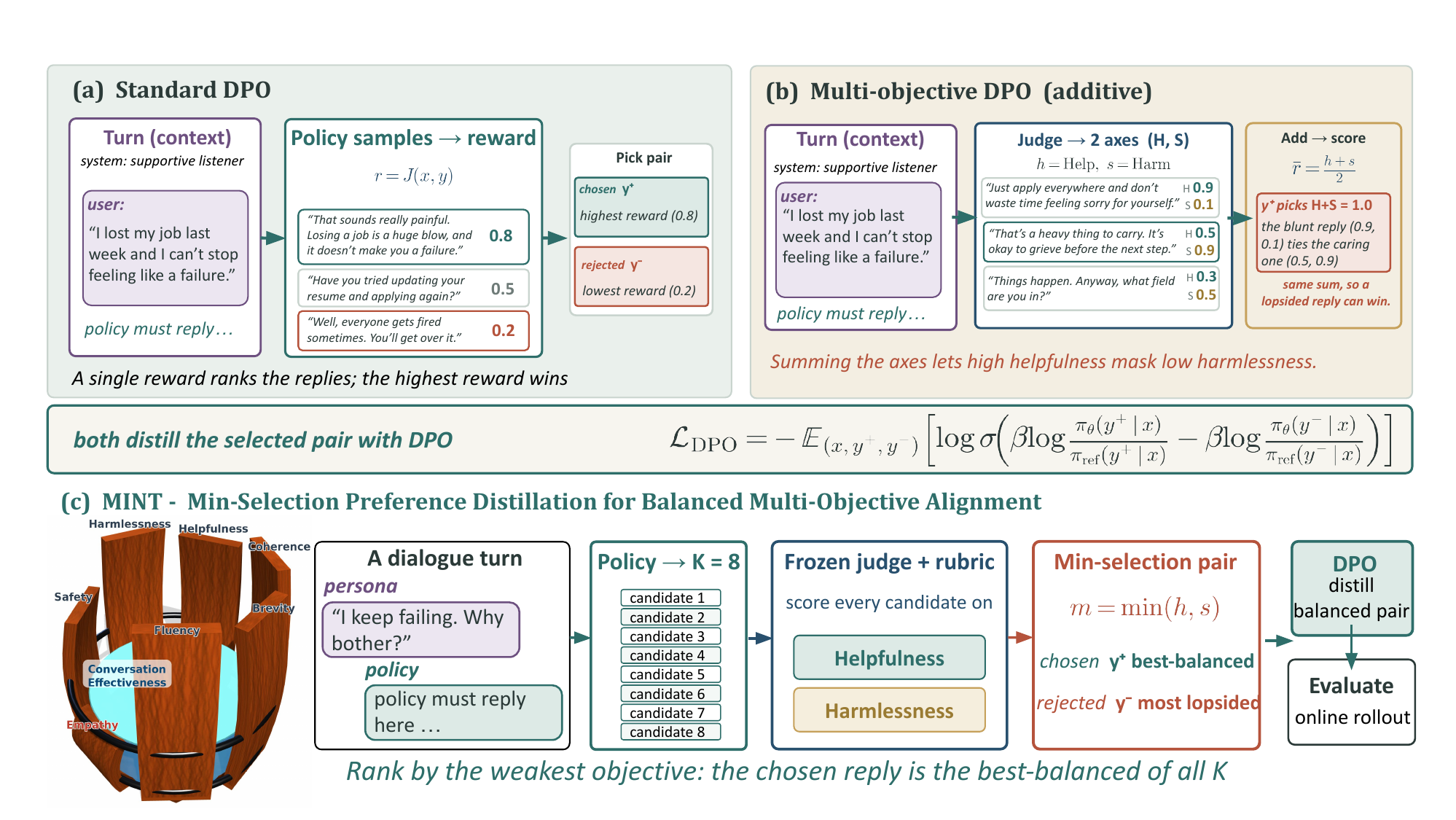}
  \caption{\textbf{Three ways to form preference pairs for DPO.}
  Every method turns one dialogue turn into a chosen/rejected pair and
  distills it with the same DPO objective; they differ only in how that
  pair is selected.
  \textbf{(a)~Standard DPO} scores each candidate reply with a single
  scalar reward $r=J(x,y)$ and takes the highest as $y^{+}$ and the
  lowest as $y^{-}$.
  \textbf{(b)~Additive multi-objective DPO} scores each candidate on two
  axes, helpfulness ($h$) and harmlessness ($s$), and ranks by their sum
  $\bar r=\tfrac{h+s}{2}$; because a surplus on one axis compensates a
  deficit on the other, a lopsided reply can be selected.
  \textbf{(c)~\textsc{Mint} (ours)} samples $K{=}8$ candidates per turn,
  scores each on both axes with a frozen judge, and selects the pair by
  the \emph{weakest} objective, $m=\min(h,s)$: the chosen reply is the
  best-balanced candidate and the rejected reply the most lopsided one,
  so both objectives improve together. Like a barrel whose water level
  is capped by its shortest stave, overall conversation effectiveness is
  bounded by the weakest objective.}
  \label{fig:overview}
\end{figure*}

\textsc{Mint} is a preference-distillation pipeline that trains a policy
to satisfy several objectives \emph{simultaneously}. At each turn it
samples a set of candidate responses from a reference policy, scores
every candidate on each objective with a frozen judge, selects a
chosen/rejected pair according to a balance-seeking rule, and distills
the resulting preferences into the policy with Direct Preference
Optimization (DPO). The only departure from a standard best-of-$K$
DPO pipeline is the \emph{selection rule} that forms the pairs; this is
where multi-objective balance is enforced.

\subsection{Problem setup}
\label{sec:setup}

We consider a turn-level dialogue policy $\pi_\theta$ that, given a
conversation prefix $x$ (system prompt and dialogue history), produces
the next agent utterance $y \sim \pi_\theta(\cdot \mid x)$. Response
quality is not scalar: we assume $m$ interpretable objectives, each
scored by a frozen judge $J$ on $[0,1]$. In our instantiation $m{=}2$,
with an \emph{action quotient} $\mathrm{AQ}(x,y)$ measuring task
progress and an \emph{emotional quotient} $\mathrm{EQ}(x,y)$ measuring
relational quality, but the method is agnostic to $m$ and to the
semantics of the objectives. The judge is a frozen large language model
prompted with an objective-specific rubric; it is never fine-tuned, so
the same backbone scores all policies and both objectives, differing
only in the rubric it is given.

The central difficulty is that the objectives are in tension: a response
can advance the task while alienating the user, or comfort the user
while stalling progress. A policy trained to maximize a naive
combination of the two, most commonly their sum, can improve the
aggregate while \emph{degrading} the weaker objective, collapsing onto
whichever axis is easier to raise. \textsc{Mint} addresses this at the
level of preference selection.

\subsection{Candidate generation and scoring}
\label{sec:candidates}

For each agent turn in a corpus of $N$ conversations, we draw $K$
candidate responses $\{y_1,\dots,y_K\}$ from a reference policy
$\pi_{\text{ref}}$ and score each on every objective, yielding reward
vectors $\mathbf{r}_k = (\mathrm{AQ}(x,y_k), \mathrm{EQ}(x,y_k))$. To
obtain candidates that genuinely span the objective trade-off, rather
than $K$ near-duplicates, we condition generation on a small set of
lightweight \emph{stance} prompts that steer the sampler toward
different regions of objective space (e.g.\ task-focused vs.\
rapport-focused); the stance is used only to diversify candidates and is
discarded afterwards, so the distilled policy is never conditioned on
it. All candidates, their reward vectors, and the original human turn
are cached, so that preference pairs under \emph{any} selection rule can
be constructed offline without re-running the judge.

\subsection{Generalized-mean scalarization of multiple objectives}
Let a turn receive two reward scores, $a=\mathrm{AQ}\in[0,1]$ (action
quotient) and $b=\mathrm{EQ}\in[0,1]$ (emotional quotient). Ranking
candidates requires collapsing the reward vector $(a,b)$ into a scalar,
for which we use the \textbf{generalized (power) mean}
\begin{equation}
M_p(a,b) = \left(\tfrac{1}{2}\big(a^{p}+b^{p}\big)\right)^{1/p},
\qquad p\in\mathbb{R}.
\label{eq:power-mean}
\end{equation}
The single parameter $p$ controls how strongly imbalance is penalized, and
recovers familiar means as special cases: the arithmetic mean
($p{=}1$), quadratic ($p{=}2$), geometric ($p{\to}0$, i.e.\ $\sqrt{ab}$),
harmonic ($p{=}{-}1$), and, in the limit, $M_{-\infty}(a,b)=\min(a,b)$.
As $p$ decreases the score is increasingly dominated by the \emph{smaller}
coordinate. At $p{=}1$ a shortfall on one axis is fully compensable by a
surplus on the other, so a lopsided turn can score as well as a balanced
one; $p{=}2$ is worse still, rewarding vector \emph{magnitude} and thus
favoring single-axis extremes. Below $p{=}1$ the ordering reverses:
the geometric mean collapses toward $0$ if either objective is near $0$,
the harmonic mean penalizes imbalance more sharply, and $M_{-\infty}$
ignores any surplus on the stronger axis entirely, scoring each candidate
purely by its weakest objective. This last limit is the selection rule
\textsc{Mint} adopts.

\subsection{Min-selection preference pairs}
\label{sec:pairs}

Given the scalarization $M_p$, we form a single preference pair per turn.
The \emph{chosen} response is the candidate that maximizes $M_p$ and the
\emph{rejected} response the one that minimizes it:
\begin{equation}
y^{+} = \arg\max_{k} M_p(\mathbf{r}_k), \qquad
y^{-} = \arg\min_{k} M_p(\mathbf{r}_k).
\label{eq:select}
\end{equation}
Throughout this work we use the limiting case $p \to -\infty$, i.e.\
$M_{-\infty} = \min(\mathrm{AQ},\mathrm{EQ})$, which we refer to as
\emph{min-selection}. Under min-selection the chosen response is the
candidate whose \emph{weaker} objective is strongest, the best-balanced
option, and the rejected response is the most lopsided one. Because the
min is insensitive to any surplus on the stronger axis, ties are common
(several candidates may share the same minimum coordinate); we break
them by the sum, i.e.\ we rank candidates by the ordered key
$\big(\min(a,b),\, a+b\big)$, so that among equally-balanced candidates
the one with greater total reward is preferred. To ensure each pair
carries a meaningful training signal, we discard turns whose chosen and
rejected candidates differ by less than a margin $\delta$ in
$\min(\mathrm{AQ},\mathrm{EQ})$; such turns are typically openers or
purely factual exchanges on which all candidates behave alike.

\subsection{Preference distillation}
\label{sec:dpo}

The selected pairs $\{(x, y^{+}, y^{-})\}$ are distilled into the policy
with DPO \citep{rafailov2023direct}, which optimizes
\begin{equation}
\mathcal{L}_{\text{DPO}} =
-\,\mathbb{E}\Big[
\log \sigma\!\big(
\beta \log \tfrac{\pi_\theta(y^{+}\mid x)}{\pi_{\text{ref}}(y^{+}\mid x)}
- \beta \log \tfrac{\pi_\theta(y^{-}\mid x)}{\pi_{\text{ref}}(y^{-}\mid x)}
\big)\Big],
\label{eq:dpo}
\end{equation}
with reference policy $\pi_{\text{ref}}$ and temperature $\beta$. We
take $\pi_{\text{ref}}$ to be the same policy used to generate the
candidates: a supervised (SFT) checkpoint where one is available, and
the base model otherwise, and train a lightweight low-rank (LoRA)
adapter on top, so that distillation adjusts the policy relative to its
own sampling distribution. This is a single \emph{offline} round:
candidates are generated once from $\pi_{\text{ref}}$, scored once, and
distilled once. The generalized-mean formulation
(Eq.~\ref{eq:power-mean}) also admits a \emph{semi-online} variant in
which candidates are regenerated from the improved policy over several
rounds; we present the general algorithm below and use the single-round
($R{=}1$) instantiation in all experiments.

\subsection{Algorithm}
\label{sec:algorithm}

\begin{algorithm}[t]
\caption{\textsc{Mint}: preference distillation with $p$-norm multi-objective selection}
\label{alg:mint}
\begin{algorithmic}[1]
\Require reference policy $\pi_{\text{ref}}$; frozen judge $J$;
  conversations $\mathcal{D}$; candidates per turn $K$; norm $p$; margin $\delta$; rounds $R$
\State $\pi_\theta \gets \pi_{\text{ref}}$
\For{round $= 1$ to $R$}
  \State $\mathcal{P} \gets \varnothing$
  \For{each conversation in $\mathcal{D}$, each agent turn with prefix $x$}
    \State sample candidates $y_1,\dots,y_K \sim \pi_\theta(\cdot \mid x)$
    \State score $\mathbf{r}_k \gets \big(\mathrm{AQ}(x,y_k),\, \mathrm{EQ}(x,y_k)\big)$ with $J$, for all $k$
    \State $y^{+} \gets \arg\max_k M_p(\mathbf{r}_k)$; \quad
           $y^{-} \gets \arg\min_k M_p(\mathbf{r}_k)$
           \Comment{ties broken by $a+b$}
    \If{$M_p(\mathbf{r}^{+}) - M_p(\mathbf{r}^{-}) \ge \delta$}
      \State $\mathcal{P} \gets \mathcal{P} \cup \{(x, y^{+}, y^{-})\}$
    \EndIf
  \EndFor
  \State $\pi_\theta \gets \arg\min_\theta \mathcal{L}_{\text{DPO}}(\pi_\theta, \pi_{\text{ref}}, \mathcal{P})$
\EndFor
\State \Return $\pi_\theta$
\end{algorithmic}
\end{algorithm}

Algorithm~\ref{alg:mint} is written in full generality: any norm $p$
(hence any point on the additive--geometric--min spectrum), any number
of rounds $R$, and any number of objectives $m$ (the reward vector
$\mathbf{r}_k$ and $M_p$ extend to $m$ coordinates unchanged). Our
experiments use the balance-maximizing extreme $p \to -\infty$ (min),
a single offline round $R{=}1$, and $m{=}2$ objectives.
\section{Results}
\label{sec:results}

We evaluate \textsc{Mint} on two structurally distinct dialogue domains
using two complementary protocols. \textbf{ESConv}
\citep{liu2021towards} is a cooperative emotional-support corpus;
\textbf{CraigslistBargain} \citep{he2018decoupling} is an adversarial
buyer--seller price negotiation. For each domain we distill from $300$
conversations of best-of-$K$ candidates ($K{=}8$) and hold out a
$10\%$ test split of \emph{conversations} that is never used for
training or checkpoint selection, so all reported numbers are on
dialogues unseen during distillation. Both objectives: achievement
quotient (AQ) and emotional quotient (EQ), are scored by the same
frozen Llama-3.1-8B judge, differing only in the domain-specific rubric.
We report each objective, their minimum $\min(\text{AQ},\text{EQ})$
(the quantity \textsc{Mint} selects on), and their absolute imbalance
$|\text{AQ}-\text{EQ}|$.

\subsection{Static held-out evaluation}
\label{sec:static-eval}

Our primary protocol is a matched, single-turn completion task on
held-out test conversations. At every real agent turn, each policy
generates one response from the \emph{identical} human-written
conversation prefix, and the judge scores all policies plus the
original human turn against the same context. This isolates the effect
of the policy while holding dialogue history fixed, avoiding the
distributional drift of full rollouts.

\begin{figure*}[t]
\centering
\includegraphics[width=\textwidth]{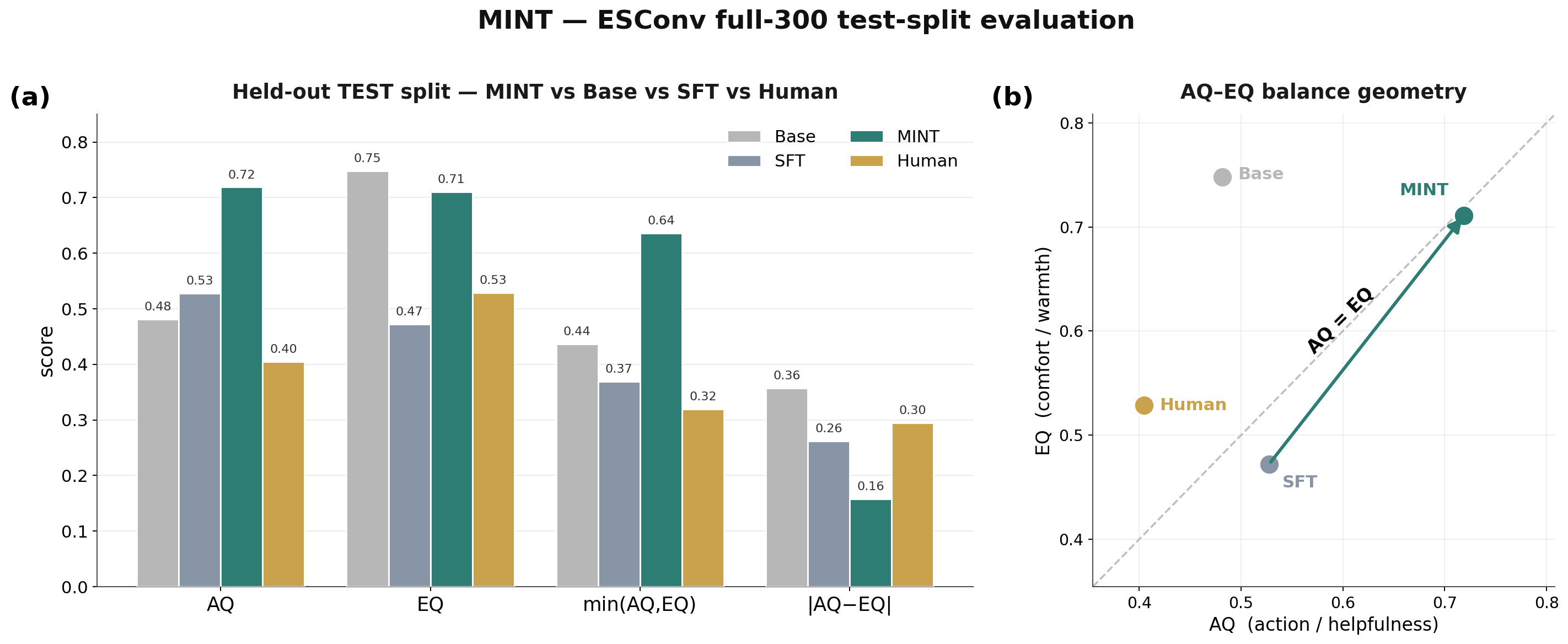}
\caption{\textbf{\textsc{Mint} on ESConv (static held-out evaluation).}
Both panels report the held-out test split ($n{=}406$ agent turns across
$30$ conversations never seen during distillation or checkpoint
selection); achievement (AQ) and emotional (EQ) quality are each scored
by a frozen Llama-3.1-8B judge.
\textbf{(a)} Per-objective scores for the base model, the supervised
(SFT) baseline, \textsc{Mint}, and human supporters. \textsc{Mint}
improves \emph{both} objectives and their minimum
$\min(\text{AQ},\text{EQ})$ over every baseline while shrinking the
imbalance $|\text{AQ}-\text{EQ}|$: relative to SFT it raises
$\min(\text{AQ},\text{EQ})$ from $0.37$ to $0.64$ (a $+0.27$ paired
improvement; Wilcoxon signed-rank $p \approx 3\times10^{-43}$, higher on
$79\%$ of matched turns).
\textbf{(b)} The same policies as points in objective space. The base
model is warm but low-achievement (high EQ, low AQ), SFT is under-warm,
and human supporters are themselves lopsided; \textsc{Mint} is the only
policy that moves toward the balance diagonal ($\text{AQ}{=}\text{EQ}$),
attaining both the highest and the most balanced scores, more balanced
than the human experts whose transcripts seed the pipeline. All scores
are quality \emph{as assessed by the reward model}}
\label{fig:esconv-static}
\end{figure*}

\begin{figure*}[t]
\centering
\includegraphics[width=\textwidth]{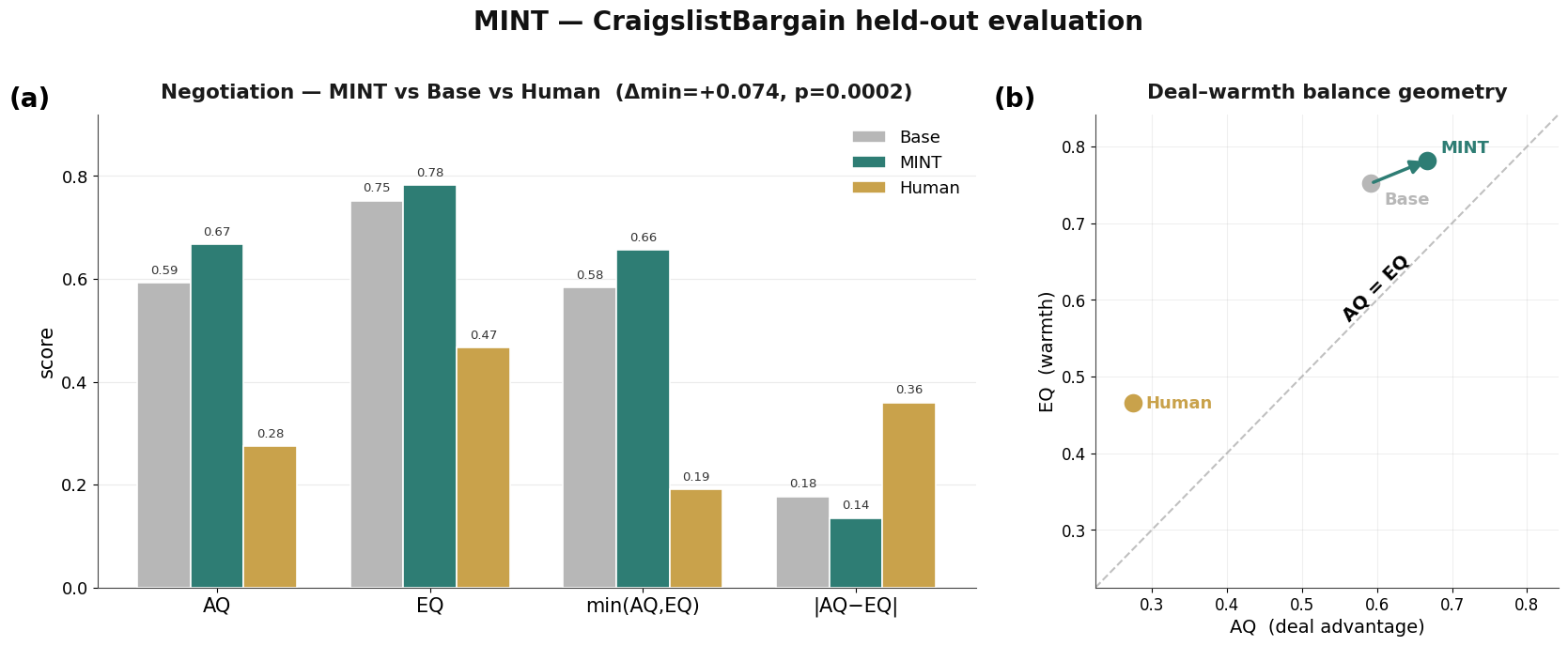}
\caption{\textbf{\textsc{Mint} on Craigslist Bargaining (held-out evaluation).}
Each policy is scored by the frozen Llama-3.1-8B judge on held-out
negotiation contexts ($n{=}103$ turns).
\textbf{(a)} Per-objective scores. \textsc{Mint} (DPO) improves both
objectives and their minimum $\min(\text{AQ},\text{EQ})$ over the base
model, raising the minimum from $0.583$ to $0.657$
($\Delta_{\min}{=}0.074$, $p{=}0.0002$) while also tightening the gap
between objectives ($0.177$ to $0.135$).
\textbf{(b)} The same policies as points in objective space. The
base$\to$\textsc{Mint} shift moves up and toward the balance diagonal
($\text{AQ}{=}\text{EQ}$). The human row is scored on real conversation
contexts and sits well below both policies on quality as assessed by the
reward model (AQ $0.275$, EQ $0.466$), serving as an \emph{unmatched}
reference. All scores are quality \emph{as assessed by the reward
model}.}
\label{fig:craigslist-heldout}
\end{figure*}


\begin{table*}[t]
\centering
\renewcommand{\arraystretch}{1.15}
\setlength{\tabcolsep}{5pt}
\small
\begin{tabular}{l cccc cccc ccc}
\toprule
& \multicolumn{4}{c}{\textbf{Reference}} & \multicolumn{4}{c}{\textbf{\textsc{Mint}}} & \multicolumn{3}{c}{\textbf{\textsc{Mint} $-$ Ref}} \\
\cmidrule(lr){2-5}\cmidrule(lr){6-9}\cmidrule(lr){10-12}
\textbf{Depth} & AQ & EQ & $|\Delta|^{\dagger}$ & min$^{\ddagger}$ & AQ & EQ & $|\Delta|^{\dagger}$ & min$^{\ddagger}$ & $\Delta$AQ & $\Delta$EQ & $\Delta$min \\
\midrule
\multicolumn{12}{l}{\cellcolor{gray!15}\textbf{ESConv} (emotional support): \textsc{Mint} vs.\ SFT reference -- advantage persists across depth} \\
\midrule
$\leq$1  & 0.356 & 0.623 & 0.380 & 0.300 & 0.535 & 0.777 & 0.242 & 0.535 & \cellcolor{teal!20}+0.179 & \cellcolor{teal!20}+0.154 & \cellcolor{teal!30}\textbf{+0.235} \\
$\leq$2  & 0.472 & 0.644 & 0.287 & 0.415 & 0.643 & 0.788 & 0.171 & 0.630 & \cellcolor{teal!20}+0.171 & \cellcolor{teal!20}+0.144 & \cellcolor{teal!30}\textbf{+0.215} \\
$\leq$3  & 0.519 & 0.568 & 0.282 & 0.403 & 0.695 & 0.772 & 0.158 & 0.655 & \cellcolor{teal!20}+0.176 & \cellcolor{teal!20}+0.204 & \cellcolor{teal!30}\textbf{+0.252} \\
$\leq$4  & 0.498 & 0.485 & 0.299 & 0.342 & 0.708 & 0.692 & 0.207 & 0.596 & \cellcolor{teal!20}+0.210 & \cellcolor{teal!20}+0.207 & \cellcolor{teal!30}\textbf{+0.254} \\
$\leq$5  & 0.457 & 0.393 & 0.302 & 0.274 & 0.683 & 0.605 & 0.240 & 0.524 & \cellcolor{teal!20}+0.226 & \cellcolor{teal!20}+0.212 & \cellcolor{teal!30}\textbf{+0.250} \\
$\leq$6  & 0.408 & 0.332 & 0.282 & 0.229 & 0.637 & 0.527 & 0.251 & 0.456 & \cellcolor{teal!20}+0.229 & \cellcolor{teal!20}+0.195 & \cellcolor{teal!30}\textbf{+0.227} \\
$\leq$7  & 0.367 & 0.291 & 0.256 & 0.201 & 0.592 & 0.465 & 0.260 & 0.399 & \cellcolor{teal!20}+0.225 & \cellcolor{teal!20}+0.174 & \cellcolor{teal!30}\textbf{+0.198} \\
$\leq$8  & 0.325 & 0.261 & 0.228 & 0.179 & 0.548 & 0.419 & 0.254 & 0.357 & \cellcolor{teal!20}+0.223 & \cellcolor{teal!20}+0.158 & \cellcolor{teal!30}\textbf{+0.178} \\
$\leq$9  & 0.295 & 0.241 & 0.212 & 0.162 & 0.508 & 0.384 & 0.249 & 0.322 & \cellcolor{teal!20}+0.213 & \cellcolor{teal!20}+0.143 & \cellcolor{teal!30}\textbf{+0.159} \\
$\leq$10 & 0.268 & 0.222 & 0.196 & 0.147 & 0.471 & 0.353 & 0.239 & 0.292 & \cellcolor{teal!20}+0.203 & \cellcolor{teal!20}+0.131 & \cellcolor{teal!30}\textbf{+0.145} \\
\midrule
\multicolumn{12}{l}{\cellcolor{gray!15}\textbf{CraigslistBargain} (negotiation): \textsc{Mint} vs.\ base reference -- advantage decays with depth} \\
\midrule
$\leq$1  & 0.507 & 0.693 & 0.218 & 0.491 & 0.575 & 0.749 & 0.188 & 0.568 & \cellcolor{teal!20}+0.068 & \cellcolor{teal!20}+0.056 & \cellcolor{teal!30}\textbf{+0.077} \\
$\leq$2  & 0.606 & 0.738 & 0.156 & 0.594 & 0.642 & 0.747 & 0.132 & 0.629 & +0.036 & +0.009 & \cellcolor{teal!12}+0.035 \\
$\leq$3  & 0.642 & 0.749 & 0.136 & 0.628 & 0.667 & 0.757 & 0.118 & 0.653 & +0.024 & +0.008 & +0.025 \\
$\leq$4  & 0.626 & 0.759 & 0.164 & 0.611 & 0.649 & 0.767 & 0.149 & 0.634 & +0.023 & +0.008 & +0.023 \\
$\leq$5  & 0.592 & 0.767 & 0.205 & 0.577 & 0.607 & 0.774 & 0.196 & 0.593 & +0.015 & +0.006 & +0.015 \\
$\leq$6  & 0.550 & 0.772 & 0.249 & 0.537 & 0.558 & 0.778 & 0.245 & 0.546 & +0.008 & +0.006 & +0.009 \\
$\leq$7  & 0.525 & 0.775 & 0.274 & 0.513 & 0.532 & 0.780 & 0.271 & 0.521 & +0.007 & +0.006 & +0.008 \\
$\leq$8  & 0.508 & 0.776 & 0.292 & 0.496 & 0.507 & 0.781 & 0.296 & 0.496 & +0.000 & +0.006 & +0.001 \\
\bottomrule
\end{tabular}
\caption{\textbf{Advantage of \textsc{Mint} by conversation depth in
interactive self-play rollouts.} Each row aggregates turns $1$ to $K$
across $30$ conversations; AQ and EQ are the mean achievement and
emotional rewards over those turns.
$^{\dagger}|\Delta|=\operatorname{mean}_i|\mathrm{AQ}_i-\mathrm{EQ}_i|$ is
the mean \emph{per-turn} imbalance (computed turn-by-turn, then averaged;
it therefore does \emph{not} equal $|\overline{\mathrm{AQ}}-\overline{\mathrm{EQ}}|$).
$^{\ddagger}\text{min}=\operatorname{mean}_i\min(\mathrm{AQ}_i,\mathrm{EQ}_i)$
is the mean \emph{per-turn} minimum (likewise computed per turn, so it is
not $\min(\overline{\mathrm{AQ}},\overline{\mathrm{EQ}})$). The last three
columns are \textsc{Mint} minus the reference on each quantity.
\textbf{Top (ESConv):} against the imbalanced SFT reference, gains are
large and \emph{persist} at all depths (all $p<10^{-4}$); SFT degrades
over multi-turn dialogue while \textsc{Mint} stays balanced.
\textbf{Bottom (Craigslist):} against a competent base, \textsc{Mint}
significantly improves the opening matched turn ($\Delta$min$=+0.077$,
$p=0.016$) but the advantage \emph{decays} as both converge. $\Delta$min
shading marks significance (dark: $p<0.05$).}
\label{tab:depth}
\end{table*}

\paragraph{Emotional support (ESConv).}
Figure~\ref{fig:esconv-static} reports results on the $30$ held-out
test conversations ($n{=}406$ agent turns). Relative to the supervised
(SFT) baseline, \textsc{Mint} improves \emph{both} objectives
simultaneously: AQ from $0.53$ to $0.72$ and EQ from $0.47$ to
$0.71$m and raises the weaker-axis reward $\min(\text{AQ},\text{EQ})$
from $0.37$ to $0.64$, a paired improvement of $+0.27$ that is highly
significant (Wilcoxon signed-rank $p \approx 3\times10^{-43}$;
\textsc{Mint} scores higher on $79\%$ of matched turns). At the same
time the imbalance $|\text{AQ}-\text{EQ}|$ falls from $0.26$ to $0.16$.
The balance geometry (Figure~\ref{fig:esconv-static}, right) makes the
mechanism visible: the base model is warm but low-achievement (high EQ,
low AQ), SFT is under-warm, and human supporters are themselves
lopsided, whereas \textsc{Mint} is the only policy that moves toward the
$\text{AQ}{=}\text{EQ}$ diagonal, attaining both the highest and the
most balanced scores, more balanced, in fact, than the human experts
whose data seeded the pipeline.

\paragraph{Negotiation (CraigslistBargain).}
Figure~\ref{fig:craigslist-heldout} reports the same protocol on the
held-out negotiation test split, with one deliberate difference from the
ESConv setup: here we omit the SFT stage entirely and apply DPO directly
on top of the prompted base model. This tests two things. First, whether
the balancing effect survives without a masked-language-modeling SFT
stage that first teaches the policy to imitate reference seller turns,
i.e., whether DPO alone can induce the improvement rather than merely
sharpening a distribution already shaped by SFT. Second, whether
\textsc{Mint} outperforms the base model it is trained from, without the
confound of an intermediate SFT checkpoint. Both hold: as in the
cooperative domain, \textsc{Mint} improves the weaker objective and
reduces $|\text{AQ}-\text{EQ}|$ relative to the base policy
(DPO improves $\min(\text{AQ},\text{EQ})$ from $0.583$ to $0.657$;
Wilcoxon $p = 0.0002$; $n = 103$ turns), while tightening the objective
gap from $0.177$ to $0.135$. This confirms that the balancing effect
transfers to an adversarial setting and a different policy model, does
not depend on a prior SFT stage, and is not an artifact of the
cooperative, emotionally-toned ESConv distribution.

\subsection{Interactive rollout evaluation}
\label{sec:rollout-eval}

The static protocol scores single completions on human-written
contexts. To test whether balance persists when a policy drives an
entire conversation, compounding its own decisions turn over turn, we
additionally run an interactive rollout protocol. Each policy conducts a
full multi-turn dialogue against a simulated conversational partner (a
persona-conditioned user model for ESConv; a prompted counterpart-agent
for CraigslistBargain), and the judge scores every policy turn along the
generated trajectory. This is a harder and noisier test, since errors
accumulate and the policy leaves the distribution of human dialogue
histories.

\paragraph{Emotional support (ESConv).}
Each policy conducts a full support dialogue against a persona-conditioned
user model, and the frozen Llama-3.1-8B judge scores every policy turn
($n{\approx}280$ turns across $30$ rollouts per policy). Against the SFT
reference from which it was distilled, \textsc{Mint}'s advantage is large
and \emph{persists at every depth}: the improvement in
$\min(\text{AQ},\text{EQ})$ ranges from $+0.235$ at the opening turn to
$+0.145$ over full ten-turn rollouts, significant throughout (Wilcoxon
$p<10^{-4}$; Table~\ref{tab:depth}, top). The mechanism is visible in the
trajectory: the SFT policy \emph{degrades} as the dialogue lengthens, its
mean minimum falling monotonically from $0.300$ to $0.147$ as early warmth
gives way to terse, low-empathy turns (EQ $0.623{\to}0.222$), whereas
\textsc{Mint} decays gracefully ($0.535{\to}0.292$) and stays balanced. In
a degenerate reference, errors compound over the trajectory, so
\textsc{Mint}'s per-turn edge accumulates rather than washes out.

Against the \emph{base} model the picture is more nuanced. The base is not
degenerate but lopsided: persistently warm (EQ near $0.80$ throughout)
while contributing little achievement early. \textsc{Mint} starts far
ahead ($\Delta\min={+}0.328$, $p{=}0.0004$, as the base's cold-start AQ of
$0.21$ drags down its minimum), but the gap closes as the base's
achievement rises, reaching parity by turn~5 ($\Delta\min={-}0.001$) and a
small non-significant reversal after. The base's stubborn warmth lifts its
running minimum to meet \textsc{Mint}'s, which trades warmth for
achievement as it presses the task. \textsc{Mint} thus dominates a
\emph{collapsing} reference at all depths but holds only an early-turn
edge over a warm-but-passive one, a distinction the pooled average ($\min$:
base $0.315$, \textsc{Mint} $0.292$) obscures entirely. The human row,
scored on real contexts rather than the rollouts, is an \emph{unmatched}
reference.

\paragraph{Negotiation (CraigslistBargain).}
Each seller policy negotiates against a prompted buyer agent (base-Gemma
conditioned on a buyer persona) seeded from real Craigslist openings, with
the judge scoring every seller turn. As the adversarial analogue of ESConv,
the counterpart actively pushes the price down, penalising a policy that
has merely learned to sound agreeable. The result mirrors \textsc{Mint}'s
behaviour against the ESConv \emph{base} rather than SFT. On the opening,
strictly matched turn, \textsc{Mint} significantly improves the weaker axis
($\Delta\min={+}0.077$, $p{=}0.016$; AQ $0.507{\to}0.575$, EQ
$0.693{\to}0.749$), reproducing the static held-out result. But the
advantage decays monotonically with depth ($+0.077, +0.035, \dots, +0.001$
from turn~1 to~8), significant only at the first exchange; by the rollout's
end the policies are indistinguishable ($\Delta\min={+}0.001$, $p{=}0.43$
pooled). The cause is not weak negotiation but a strong reference: the base
Gemma-3-27B seller is \emph{already} balanced under our prompt, holding
price, justifying value, and staying courteous (per-turn minimum $0.49$ at
the opening and rising). With little imbalance to correct, the two policies
converge to similar outcomes near the seller's target, and their per-turn
rewards equalise as the shared-opening conversations diverge into
comparably successful negotiations.

\paragraph{When does min-selection help, and for how long?}
Together the two rollouts expose a principle the static evaluations could
not reveal: \textsc{Mint} improves the weaker objective in proportion to
the reference's \emph{imbalance}, and this gain \emph{persists over
interaction only while the reference stays imbalanced along the
trajectory}. Against a badly and increasingly imbalanced reference (ESConv
SFT, whose empathy collapses over a dialogue), the advantage is large and
grows at every depth. Against a reference that is already balanced, or
imbalanced only transiently, it is significant on matched contexts (the
static evaluation and the opening turn of both rollouts) but attenuates as
trajectories diverge and a competent reference recovers the weaker axis on
its own: the ESConv base regains achievement over the conversation, and the
Craigslist base was balanced from the outset. This is why pooled averages
\emph{understate} the method, collapsing a strongly significant matched
opening effect with later turns where a competent reference has caught up,
averaging a real localised gain into an apparent null. The depth-resolved
view (Table~\ref{tab:depth}) is the honest one: min-selection is a
\emph{targeted corrector of reference imbalance}, most valuable exactly
where and when that imbalance is present, not a uniform improvement.

\subsection{Discussion}
\textsc{Mint} forms preference pairs by selecting on the \emph{minimum} of
two objectives rather than a weighted sum, yielding policies that are
stronger on both axes and better balanced between them. A weighted sum
lets a gain on the easy axis pay for a loss on the hard one; min-selection
closes that route by scoring each pair on its weaker axis. Where the
reference is badly imbalanced the effect is large and durable, raising the
weaker axis from $0.37$ to $0.64$ on emotional support and exceeding the
human supporters who seeded the distillation, which suggests the balancing
signal was latent in the best-of-$K$ candidates and surfaced only when
selection refused to average the axes away. But our turn-resolved analysis
shows the benefit tracks the reference: it compounds over interaction only
while the reference stays imbalanced, and against an already-balanced
policy it holds on matched contexts yet attenuates as trajectories diverge.
Min-selection is thus a targeted corrector of objective imbalance, most
valuable where that imbalance is most severe.
\bibliography{refs}
\clearpage
\onecolumn
\appendix
\begingroup
\etocstandardlines
\etocsettocdepth{subsection}
\etocsettocstyle{\subsection*{\centering Appendix Contents}}{}
\localtableofcontents
\endgroup
\bigskip
\section{Training Details}
\label{app:training}

\subsection{Emotional Support DPO Training}
\label{app:esconv-training}

\begin{figure}[hbp]
\centering
\includegraphics[width=\textwidth]{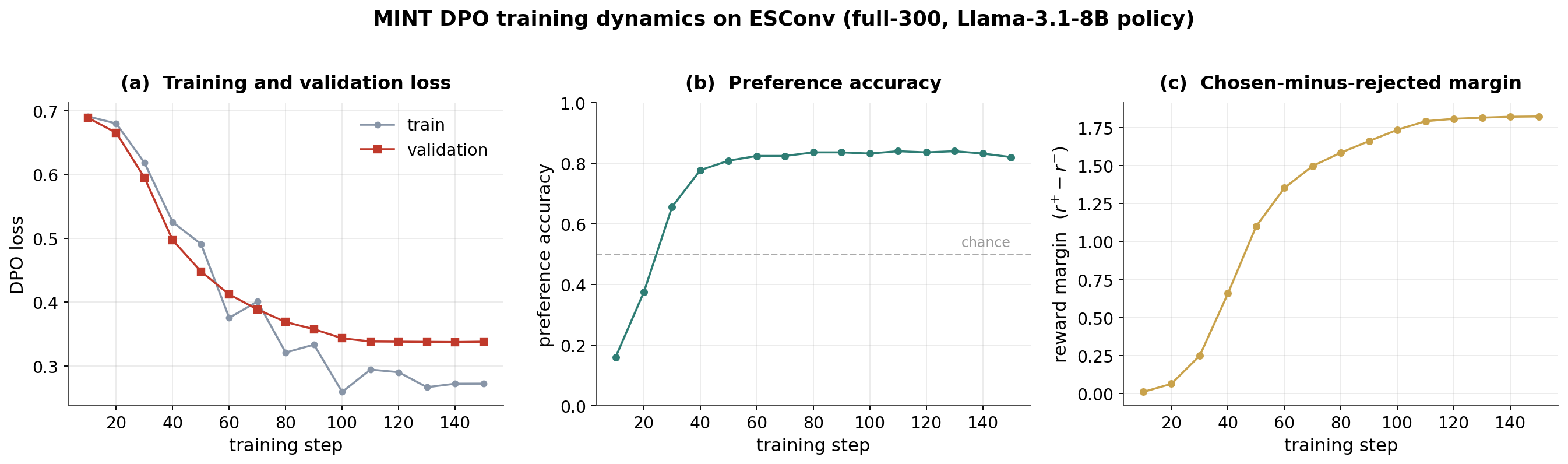}
\caption{\textbf{\textsc{Mint} training dynamics on ESConv}
(full-300 distillation, Llama-3.1-8B policy).
\textbf{(a)} Training and validation DPO loss; validation loss
decreases monotonically and flattens by roughly step $130$.
\textbf{(b)} Preference accuracy on held-out pairs, rising from
near-chance to a plateau of about $0.84$.
\textbf{(c)} Mean reward margin between the chosen and rejected
responses ($r^{+}\!-\!r^{-}$), which grows steadily without diverging.
The checkpoint with the lowest validation loss is selected under early
stopping.}
\label{fig:esconv-training}
\end{figure}

We distill the ESConv policy from a Llama-3.1-8B-Instruct backbone,
using a supervised (SFT) checkpoint as the reference policy
$\pi_{\text{ref}}$. Following a merge-then-adapt scheme, we merge the SFT
adapter into the base weights and train a fresh low-rank DPO adapter on
top (rank $16$, $\alpha{=}32$, dropout $0.05$; $41.9$M trainable
parameters, $0.52\%$ of the model), so that disabling the adapter
recovers the SFT reference exactly. Preference pairs are formed by
$\min(\text{AQ},\text{EQ})$ selection with the sum tie-break of
Section~\ref{sec:pairs} and a selection margin $\delta{=}0.25$; pairs
whose prompt or completion exceed the length budget are dropped. From
the $300$ distilled conversations this yields $1{,}956$ training and
$224$ validation pairs. We train with DPO temperature $\beta{=}0.1$,
learning rate $10^{-5}$, a cosine schedule with $10\%$ warmup, and an
effective batch size of $8$, selecting the checkpoint with the lowest
validation loss under early stopping (patience $3$). As shown in
Figure~\ref{fig:esconv-training}, validation loss falls from $0.69$ to
$0.34$, preference accuracy plateaus near $0.84$, and the
chosen/rejected reward margin grows steadily to roughly $1.8$ without
diverging, indicating stable optimization rather than reward
over-optimization.

\subsection{Price-Negotiation DPO Training}
\label{app:craigslist-training}

\begin{figure}[tbp]
\centering
\includegraphics[width=\textwidth]{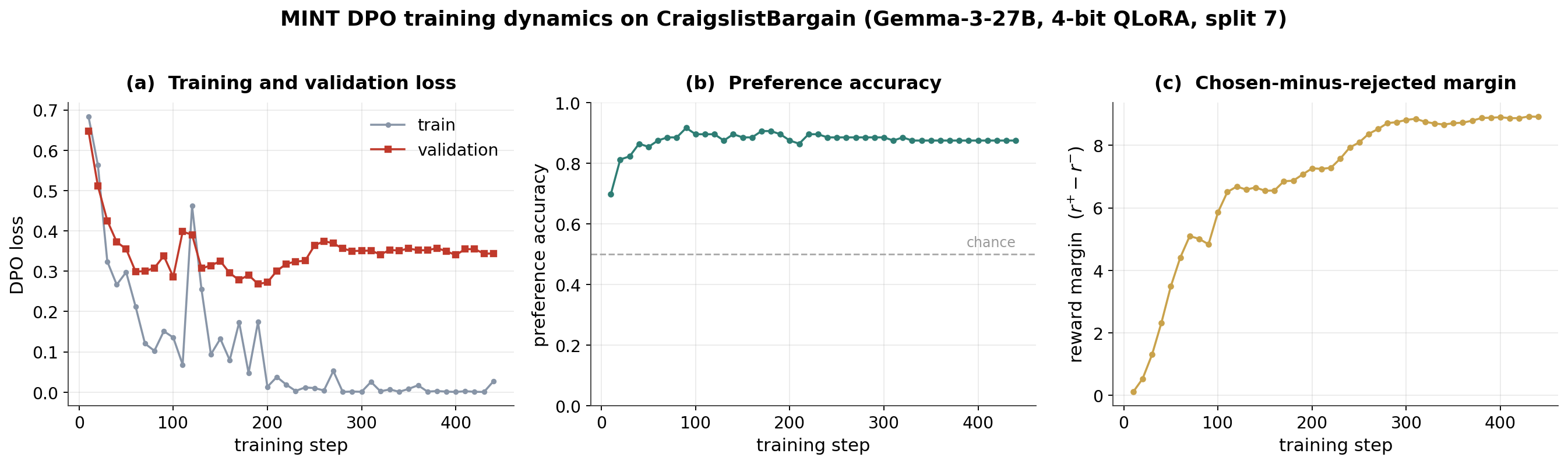}
\caption{\textbf{\textsc{Mint} training dynamics on CraigslistBargain}
(Gemma-3-27B policy, 4-bit QLoRA).
\textbf{(a)} Training and validation DPO loss; validation loss falls
sharply over the first ${\sim}50$ steps and flattens near its minimum,
while the training loss is noisier under 4-bit quantization and a small
effective batch.
\textbf{(b)} Preference accuracy on held-out pairs, rising from about
$0.76$ to a plateau near $0.84$.
\textbf{(c)} Mean reward margin between the chosen and rejected
responses ($r^{+}\!-\!r^{-}$), which grows steadily and monotonically
without diverging.
The checkpoint with the lowest validation loss is selected under early
stopping (patience $3$).}
\label{fig:craigslist-training}
\end{figure}

We distill the CraigslistBargain policy from a Gemma-3-27B-Instruct
backbone under 4-bit QLoRA (NF4 with double quantization and a bfloat16
compute dtype), using the quantized base model with its adapter disabled
as the reference policy $\pi_{\text{ref}}$, so that turning the adapter
off recovers the reference exactly. We train a single low-rank DPO
adapter on all attention and MLP projections
(\texttt{q,k,v,o,gate,up,down}) with rank $8$, $\alpha{=}16$, and dropout
$0.05$ ($56.8$M trainable parameters, $0.21\%$ of the model), using
paged 8-bit AdamW and gradient checkpointing for single-GPU memory
efficiency. Preference pairs are formed by $\min(\text{AQ},\text{EQ})$
selection with the sum tie-break of Section~\ref{sec:pairs} and a
selection margin $\delta{=}0.25$; pairs whose prompt exceeds $384$ tokens
or whose chosen/rejected completion exceeds $128$ tokens are dropped.
From the $300$ distilled conversations (each with $K{=}8$ scored
candidates per turn), split $80/10/10$ into train/validation/test, this
yields $725$ training and $89$ validation preference pairs. We train with
DPO temperature $\beta{=}0.1$, learning rate $10^{-5}$, a cosine schedule
with $10\%$ warmup, and an effective batch size of $8$ (per-device batch
$1$ with gradient accumulation $8$) for up to $5$ epochs, evaluating
every $10$ steps and selecting the checkpoint with the lowest validation
loss under early stopping (patience $3$). As shown in
Figure~\ref{fig:craigslist-training}, validation loss falls from about
$0.63$ to a minimum near $0.25$ before settling around $0.29$, preference
accuracy plateaus near $0.84$, and the chosen/rejected reward margin
grows steadily to roughly $5.8$. The larger terminal margin and noisier
training loss, relative to the ESConv run, are consistent with the more
adversarial and sharply separable preferences of price negotiation under
a 4-bit-quantized policy and a small effective batch, while the smoothly
decreasing validation loss indicates stable optimization rather than
reward over-optimization.

\section{Reward Rubrics and Judge Configuration}
\label{app:rubrics}

\subsection{Frozen judge model}
All rewards in this paper are produced by a single \emph{frozen}
\texttt{Llama-3.1-8B-Instruct} model acting as an LLM-as-judge; its
weights are never updated during rejection sampling, preference-pair
construction, or evaluation, so the reward signal is held fixed across
training and all evaluation conditions. The judge is prompted separately
for each domain with the rubric-specific system prompt given below,
receives the item/context and the conversation so far, and scores
\emph{only} the policy's most recent turn. It emits a single JSON object
containing, for each axis, a score in $[0,1]$ per criterion plus a
one-sentence rationale; decoding is greedy
($\text{temperature}=0$) for determinism. Each axis reward is the
weighted average of its criterion scores,
$r = \left(\sum_c w_c\, s_c\right) / \sum_c w_c$, with $s_c$ clamped to
$[0,1]$ and weights $w_c$ as tabulated below. The same judge, prompts,
and weighting are used for candidate scoring during training and for both
the static and rollout evaluations, so within any comparison the reward
model is identical across policies.

Two design choices are shared across domains. First, the two axes are
scored \emph{independently}: the system prompt instructs the judge not to
let one axis influence the other, and gives worked examples where the two
diverge (e.g.\ a curt but effective turn, or a warm capitulation), so
that a policy cannot inflate both axes merely by being pleasant or merely
by being effective. This orthogonality is what makes
$\min(\text{AQ},\text{EQ})$ a meaningful balance objective rather than a
proxy for overall quality. Second, each axis is decomposed into three
weighted criteria to reduce single-number judge variance and to force the
model to reason about distinct sub-qualities before aggregating.

\subsection{Emotional support (ESConv)}
For ESConv the achievement axis (AQ) measures the \emph{presence and
specificity} of actionable support, while the emotional axis (EQ)
measures \emph{validation, warmth, and timing}. Critically, premature
advice is penalised only under EQ (timing), never under AQ (substance),
so that a concrete-but-ill-timed suggestion is correctly recorded as
high-AQ/low-EQ rather than being globally downgraded. To counteract the
judge's tendency to compress AQ into a narrow band, the AQ prompt embeds
three few-shot calibration anchors spanning a bare information-gathering
question ($\approx0.1$--$0.2$), a generic untailored suggestion
($\approx0.45$--$0.55$), and a specific, tailored, actionable step
($\approx0.85$--$0.95$); the judge is instructed to interpolate between
these and to avoid assigning every action turn the same score. The
criteria and weights are given in Table~\ref{tab:rubric-esc}.

\begin{table}[h]
\centering
\small
\renewcommand{\arraystretch}{1.3}
\begin{tabular}{p{0.30\linewidth} c p{0.52\linewidth}}
\toprule
\textbf{Criterion} & \textbf{$w$} & \textbf{Description} \\
\midrule
\multicolumn{3}{l}{\cellcolor{teal!10}\textbf{AQ , Action / Helpfulness Quality} (substance and specificity)} \\
\midrule
Provides Concrete Help & 0.35 & Specificity of help: vague gesture $\approx0.2$; generic suggestion $\approx0.4$--$0.5$; specific substantive suggestion or concrete fact $\approx0.75$--$0.9$. \\
Offers a Usable Next Step & 0.30 & Vague direction $\approx0.3$; simple step $\approx0.5$; clear step tailored to the situation $\approx0.8+$. \\
Relevance to the Situation & 0.35 & Generic $\approx0.3$; relevant to the general problem $\approx0.6$; addresses specific circumstances $\approx0.9$. \\
\midrule
\multicolumn{3}{l}{\cellcolor{gold!12}\textbf{EQ , Emotional / Comfort Quality} (validation, warmth, timing)} \\
\midrule
Validates Emotions & 0.40 & Acknowledges or normalises feelings; naming/reflecting the feeling scores moderate-to-high even if brief. \\
Warm and Patient Tone & 0.30 & Calm, non-judgmental, unhurried. \\
Good Timing / Not Premature & 0.30 & Penalised \emph{only here} if advice precedes hearing the seeker out; well-timed high, premature low. \\
\bottomrule
\end{tabular}
\caption{ESConv reward rubric. AQ scores the substance and specificity of
help; EQ scores validation, warmth, and timing. Premature advice is
penalised under EQ only.}
\label{tab:rubric-esc}
\end{table}

\subsection{Negotiation (CraigslistBargain)}
For CraigslistBargain the achievement axis (AQ) measures
\emph{deal advantage for the seller},whether the turn holds or improves
the price toward the seller's target and advances a concrete
agreement,while the emotional axis (EQ) measures \emph{interpersonal
warmth of the language itself}, explicitly independent of whether the
deal is going well. The prompt is deliberately adversarial in its
framing: it states that a turn closing a good deal in a curt,
transactional manner must score \emph{low} EQ, and that a warm turn that
concedes a bad price must score \emph{high} EQ, precisely so that the two
axes cannot collapse into a single ``good negotiation'' signal. This
construction is what makes the negotiation setting a genuine test of
balance rather than of raw effectiveness. The criteria and weights are
given in Table~\ref{tab:rubric-neg}.

\begin{table}[h]
\centering
\small
\renewcommand{\arraystretch}{1.3}
\begin{tabular}{p{0.30\linewidth} c p{0.52\linewidth}}
\toprule
\textbf{Criterion} & \textbf{$w$} & \textbf{Description} \\
\midrule
\multicolumn{3}{l}{\cellcolor{teal!10}\textbf{AQ , Deal Advantage} (price/economic substance only)} \\
\midrule
Holds or Improves Price & 0.45 & Caving to a low buyer price $\approx0.1$; holding firm near target $\approx0.7$; extracting a higher price $\approx0.9$. A friendly tone does \emph{not} raise this score. \\
Justifies with Value & 0.25 & Concrete reason the price is warranted (condition, features, demand). Bare ``no'' $\approx0.2$; specific value argument $\approx0.8$. \\
Advances to Close & 0.30 & Concrete step toward finalising (counter-offer, terms). Stalling $\approx0.3$; clear actionable proposal $\approx0.8$. \\
\midrule
\multicolumn{3}{l}{\cellcolor{gold!12}\textbf{EQ , Interpersonal Warmth} (tone/language only)} \\
\midrule
Warmth of Language & 0.45 & Curt/robotic (``7, deal.'') $\approx0.1$--$0.2$ \emph{even if it closes a good deal}; warm, personable phrasing $\approx0.8+$ \emph{even if the price is bad}. \\
Acknowledges the Buyer & 0.30 & Recognises the buyer's stated situation (budget, needs). Ignoring it $\approx0.2$; empathising $\approx0.8+$. \\
Respectful \& Non-Dismissive & 0.25 & Polite, not condescending. Dismissive $\approx0.1$; respectful $\approx0.8$. Independent of deal outcome. \\
\bottomrule
\end{tabular}
\caption{CraigslistBargain reward rubric. AQ scores price/economic
substance only; EQ scores warmth of language only. The prompt explicitly
forces the axes to diverge,a curt closer scores low EQ, a warm
concession scores high AQ,so that balance cannot be achieved by raw
effectiveness alone.}
\label{tab:rubric-neg}
\end{table}

\subsection{Axis independence in practice}
The orthogonality the prompts request is borne out in the scores. On
probe turns constructed to be maximally divergent, the judge assigns the
intended opposite extremes: a curt, effective negotiation close
(``\$7, and we have a deal.'') receives high AQ but low EQ
(AQ~$\approx0.65$, EQ~$\approx0.25$), while a warm capitulation receives
the reverse. Across sampled turns the two axes are weakly correlated,
confirming that $\min(\text{AQ},\text{EQ})$ and $|\text{AQ}-\text{EQ}|$
measure balance rather than a single latent quality axis. Full system
prompts, including the JSON output schema and the ESConv calibration
anchors, are reproduced verbatim in our released code.
\end{document}